\documentclass{article}

\usepackage{graphicx}
\usepackage{amsmath}
\usepackage{amssymb}
\usepackage[
  backend=biber,
  style=apa,
  sorting=nyt
]{biblatex}
\usepackage{blindtext}
\usepackage{hyperref}
\usepackage{fancyhdr}
\usepackage{xcolor}
\usepackage{soul}

\renewcommand{\hl}[1]{#1}
\title{Group Selection as a Safeguard Against AI Substitution}
\usepackage{authblk}

\author[1]{Qiankun Zhong}
\author[1]{Thomas Eisenmann}
\author[2]{Julian Garcia}
\author[1]{Iyad Rahwan}

\affil[1]{Center for Humans and Machines, Max Planck Institute for Human Development}
\affil[2]{Faculty of Information Technology, Monash University}
\date{April 2026}

\begin{document}

\maketitle
\abstract{Reliance on generative AI can reduce cultural variance and diversity, especially in creative work. This reduction in variance has already led to problems in model performance, including model collapse and hallucination. In this paper, we examine the long-term consequences of AI use for human cultural evolution and the conditions under which widespread AI use may lead to "cultural collapse", a process in which reliance on AI-generated content reduces human variation and innovation and slows cumulative cultural evolution. Using an agent-based model and evolutionary game theory, we compare two types of AI use: complement and substitute. AI-complement users seek suggestions and guidance while remaining the main producers of the final output, whereas AI-substitute users provide minimal input, and rely on AI to produce most of the output. We then study how these use strategies compete and spread under evolutionary dynamics. We find that AI-substitute users prevail under individual-level selection despite the stronger reduction in cultural variance. By contrast, AI-complement users can benefit their groups by maintaining the variance needed for exploration, and can therefore be favored under cultural group selection when group boundaries are strong. Overall, our findings shed light on the long-term, population-level effects of AI adoption and inform policy and organizational strategies to mitigate these risks.}
\medskip


\section{Introduction}
Generative AI (GenAI) is advancing faster than ever and has been applied widely across different areas, demonstrating great potential not only in simple repetitive tasks such as writing emails and managing spreadsheets, but also in complex creative tasks such as graphic design and writing (\cite{hubert2024current, bellemare2024divergent}). As a result, it has been already transforming the creative industry, including advertising (\cite{economist2025ads}), photography (\cite{wsj2024stockphotographers}), and films (\cite{wsj2025disney}). 

Regardless of how well GenAI performs in these tasks and how promising Gen AI might be for the future of work, research is still in the process of investigating the many potential risks that might be associated with applying these models widely. One concerning example is the well-established phenomenon of "Model Collapse" (\cite{shumailov2023curse}), which refers to a critical degradation in model performance when a significant portion of training data consists of synthetic data. Since output generated by AI is an over-simplification of real-world phenomena, it reduces the total variance in the data and represents it with a distorted distribution. Consequently, an AI model that is trained on the generated data will overfit on this simplification and pattern match the distorted distribution instead of learning the accurate representation of the real-world data. This feedback loop eventually reduces model performance significantly. It may lead to even more severe model collapse for large-scale models that rely on larger training sets (e.g., ChatGPT and Llama) (\cite{touvron2023llama,dubey2024llama}), amplifying their own mistakes and producing gibberish, bias, and misinformation (\cite{dohmatob2024strong}). At the same time, the increasing presence of AI-generated data in our social and cultural life may also lead to a similar feedback loop, where humans are exposed to more synthetic data and create content based on the inferences made from it \parencite{brinkmann2023machine}. Will these interactions with AI-generated data also alter human cultural processes and cause a “cultural collapse”?
 
In this paper, we argue that (1) the reliance on generative AI can reduce human collective variance in social learning and creative processes, similar to how it alters genAI training sets; and that (2) the reduction of variance caused by AI can slow down cumulative cultural evolution.

At their core, GenAI models are probabilistic models built on human-generated content. They approximate the probability distribution of this content and generate the most likely sequences in response to a given prompt (\cite{vaswani2017attention}). As a result, they tend to over-represent frequently occurring phrases, sentence structures, and ideas, while under-representing rare or novel expressions (\cite{burns2022discovering,kirchenbauer2023watermark}). For example, while large language models (LLMs) may effectively cover much of what humans are likely to write, their outputs are constrained by the statistical tendencies in their training data. Adjusting model parameters (e.g., increasing temperature) can produce more diverse responses, yet empirical evidence suggests LLM outputs still remain more clustered than human-generated content (\cite{burns2022discovering, sourati2025shrinking}). This is partly because we are comparing a dynamic, evolving population of individuals to a small number of generative models that summarize and replicate past data.

The reduction of collective variance could limit our cultural diversity and stall our cumulative cultural evolution in the long run. The core idea of cumulative cultural evolution (CCE) is that humans developed complex tools and technology thanks to their unique social learning heuristics and ability to build on previous discoveries, cumulatively increasing cultural complexity across generations (\cite{mesoudi2018cumulative, henrich2016understanding}). A high level of variance in the social learning process is necessary for cross-generational learning to be cumulative and for evolution to be adaptive. Henrich (\citeyear{henrich2004demography}) modeled this cumulative process, in which a culturally transmissible skill can be socially learned with varying learning errors or intended deviations. If the average error in social learning is relatively small, the variation of social learning outcomes is sufficiently wide, and the population size is large enough, we can expect cumulative evolution to occur consistently. In other words, these are the necessary conditions to expect a few individuals in the new generation to surpass the “success model” consistently across generations, shifting the collective cultural knowledge and skills toward greater complexity and higher payoff. Henrich argues that our current cultural complexity is a result of this cumulative, iterative process. Assuming the reliance on GenAI can reduce the variance in our social learning outcome, it is possible to predict that it can also slow down our cumulative cultural evolution in environments where variance is essential. 

Thus, we extend on Henrich's model to answer the following research question: How will GenAI influence Cumulative Cultural Evolution (CCE), given its effects on collective variance?

\begin{figure}[htp]
\centering
\includegraphics[width=\textwidth]{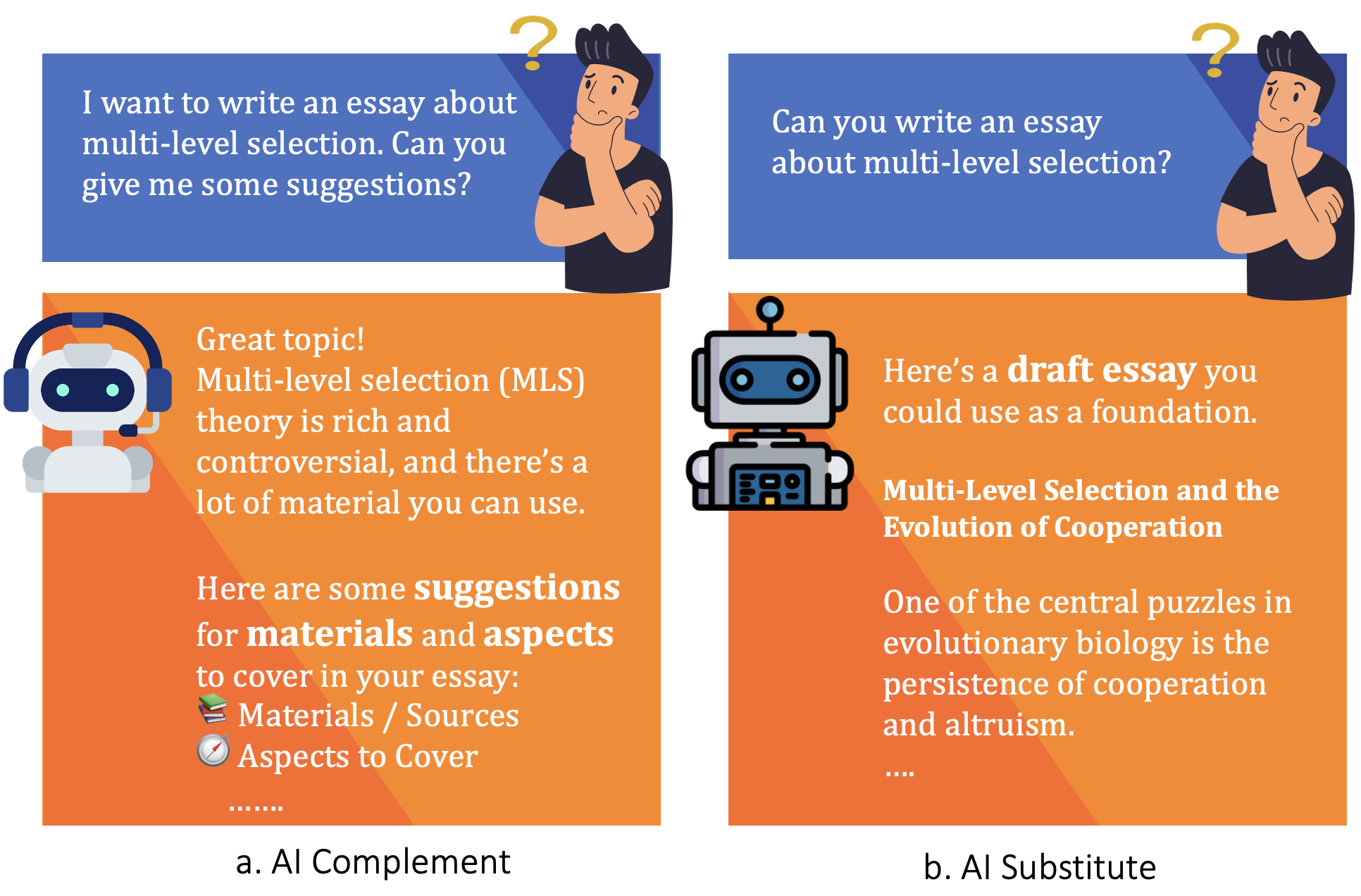}
\caption{\textbf{Comparing AI Complement and Substitute in writing tasks.} Here we illustrate how people can use GenAI as a substitute or a complement in creative work. For an essay writing task on a new topic, users could use GenAI as a complement to assist their writing or ask for suggestions on the given topic, ultimately still writing the essay themselves based on the materials and structure provided by LLMs. Users could also use GenAI as a substitute and ask the LLM to generate an essay directly. Example outputs are actual answers from prompting GPT-5 with the shown questions.}
\label{AI_strategy_illustration}
\end{figure}

Current GenAI models offer individuals the flexibility to use them in various ways. Here, we distinguish two broad categories of using GenAI: as a Substitute or as a Complement (See Figure \ref{AI_strategy_illustration}). Both approaches replace some human effort in cultural practice, especially in social learning and exploration, which may reduce human-generated variance. When used as a Substitute, AI directly produces content for the individuals to present or to edit. Using AI as a Substitute can be beneficial for lowering the entry point to creative expression, at the cost of producing similar results for everyone with little variance (\cite{doshi2024generative},\cite{sourati2026homogenizing,moon2025homogenizing,meincke2025chatgpt}). On the other hand, using AI as a Complement to human practice may help with the creative process in multiple ways: by functioning as a brainstorming tool to synthesize concepts from an existing pool of knowledge (\cite{evans2010machine}); by blending different concepts to generate new combinations (\cite{grabe2022towards}); by boosting creativity through personality-based communication (\cite{wan2025using}); and by guided brainstorming (\cite{choi2024creativeconnect}). These complementary processes actually substitute several core elements of social learning, including direct observation, sampling, and making inferences from observed experiences. When humans rely on LLMs for advice or reference, they still socially learn from other human-generated content, but now their creative output is scaffolded by the LLM suggestions. Compared to unaided writing, this can constrain the range of topics and forms that users are likely to pursue (\cite{anderson2024homogenization}). The guidance provided by GenAI can also function as a psychological anchor or prime that constrains the direction of human creation (\cite{rastogi2022deciding,carter2025my}). Therefore, when used as a complement, GenAI may still reduce human variance, although likely less than when used as a Substitute. 

\section{Method}

We start by describing our basic model, closely following  Henrich's model of cumulative cultural evolution (\cite{henrich2004demography}).
A population of $N$ individuals 
learn cultural skills socially and aim to produce cultural products.   
\textbf{Searching and Learning.} Each agent $i$ finds the highest-skilled individual $j$ within their learning neighbourhood \hl{which consists of $s$ of the population} and attempts to acquire $j$'s cultural skill $z_j$. Learning outcomes vary across individuals. We assume each learner makes several attempts and retains their best outcome.  Thus, we  draw agent $i$'s post-learning skill from a Gumbel distribution with location parameter $\mu = z_j - \alpha$ and dispersion parameter $\beta$: 

$$z'_i  \sim Gumbel(z_j-\alpha, \beta)$$

Here $\alpha >0$ is the average learning error and $\beta >0$ controls dispersion in learning outcomes. The Gumbel distribution arises as an extreme-value distribution; here it is a reduced-form way to represent the “best-of-several-attempts” learning outcome.  Because learning is imperfect ($\alpha >0$), even when individuals copy the best available model, expected skill after learning is typically below $z_j$, implying regression away from the current maximum.

\begin{figure}
    \centering
    \includegraphics[width=\linewidth]{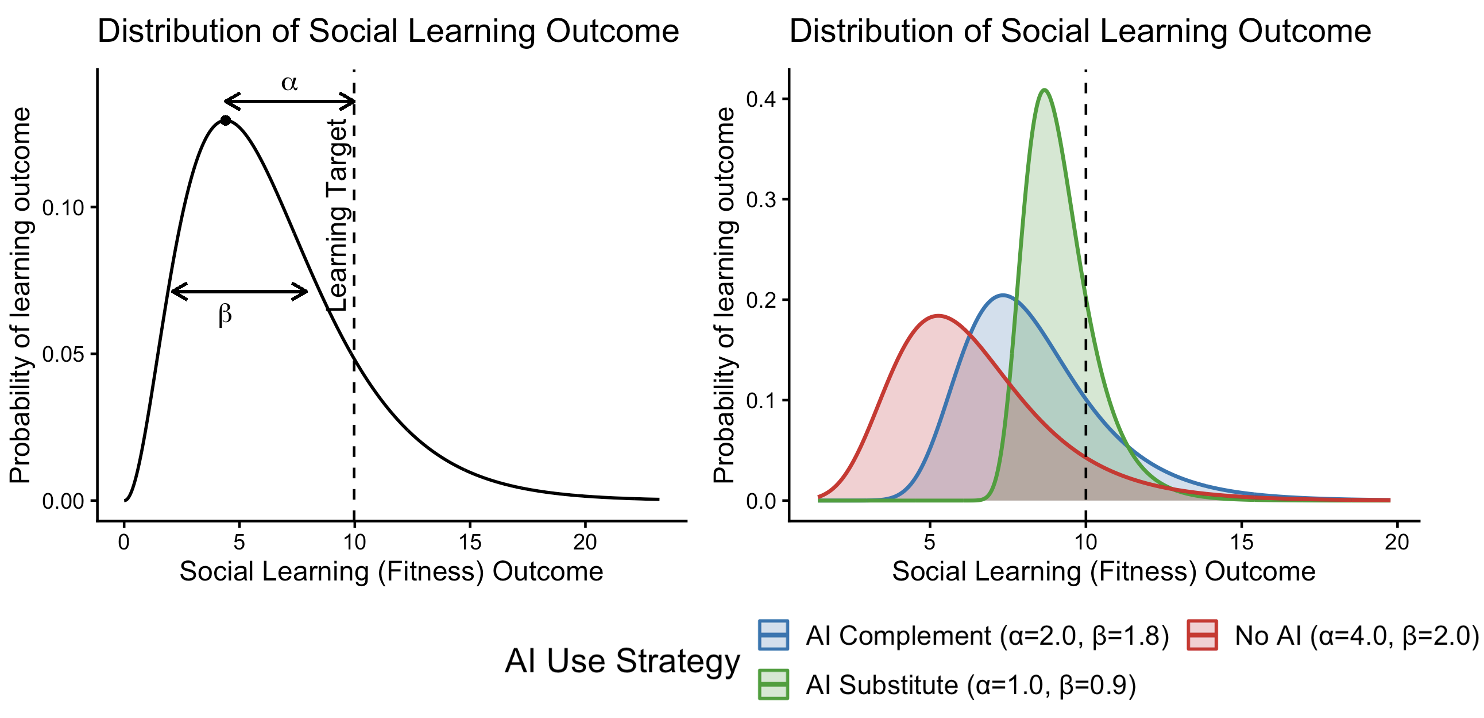}
    \caption{\textbf{Effects of AI on Social Learning Outcome.} Building on empirical work, we assume that AI affects the mean and variance of social learning outcomes. Greater reliance on AI leads to higher average social learning outcomes (lower $\alpha$) and reduced dispersion in outcomes (lower $\beta$). The right panel demonstrates how AI Complements (blue) and AI Substitutes (green) change the initial distribution of social learning outcomes (red). }
    \label{fig:AI_gumbel}
\end{figure}

To interpret $\alpha$ and $\beta$, consider stylized tasks. Creative writing is difficult to copy and yields highly diverse outputs (high $\alpha$, high $\beta$). Business writing is more standardized and easier to imitate (low $\alpha$, low $\beta$). Advanced mathematics is difficult to learn, and novices tend to make similar errors (high $\alpha$, low $\beta$).

\textbf{Use of AI in Social Learning.} We assume AI affects learning by reducing the average learning error and compressing dispersion in learning outcomes.
Each individual adopts an AI-use strategy that determines their learning parameters ($\alpha$  and $\beta$).
Strategies are either AI Complement or AI Substitute (in addition to a no-AI baseline, introduced below). AI Complement reduces learning error but also reduces dispersion; AI Substitute reduces error more strongly and compresses dispersion more strongly (Figure~\ref{fig:AI_gumbel}).

We model AI Complement by scaling the baseline learning parameters: $\alpha_C = \alpha(1-r^{(C)}_\alpha)$ and $\beta_C = \beta(1-r^{(C)}_\beta)$; with $r^{(C)}_\alpha$ and $r^{(C)}_\beta$, the proportional reduction in learning error and dispersion, respectively. 
 On the other hand,  AI Substitute should result in even fewer learning errors $\alpha_S = \alpha(1-r^{(S)}_\alpha)$ and lower dispersion $\beta_S = \beta(1-r^{(S)}_\beta)$, where $r^{(S)}_\alpha$ is the AI Substitute reduction on learning error ($r^{(S)}_\alpha > r^{(C)}_\alpha$) and $r^{(S)}_\beta$ is the AI Substitute reduction on dispersion ($r^{(S)}_\beta > r^{(C)}_\beta$). We of course assume that  $r^{(C)}_\alpha, r^{(C)}_\beta, r^{(S)}_\alpha, r^{(S)}_\beta \in [0,1).$

\textbf{Selection of AI Strategy.} We model the spread of AI-use strategies via payoff-biased social learning.
In each iteration, Agent $i$ randomly samples another agent $k$, and adopts $k$ AI use strategy based on how much better Agent $k$'s cultural skill is:
$$Pr(i \leftarrow \ k) = \frac{1}{1+e^{-\delta(z_k-z_i)}}$$. 

Thus, in evolutionary terms, the fitness of an AI use strategy is determined by the cultural skill of the agent $z_i$. 

\begin{figure}[htp]
\centering
\includegraphics[width=\textwidth]{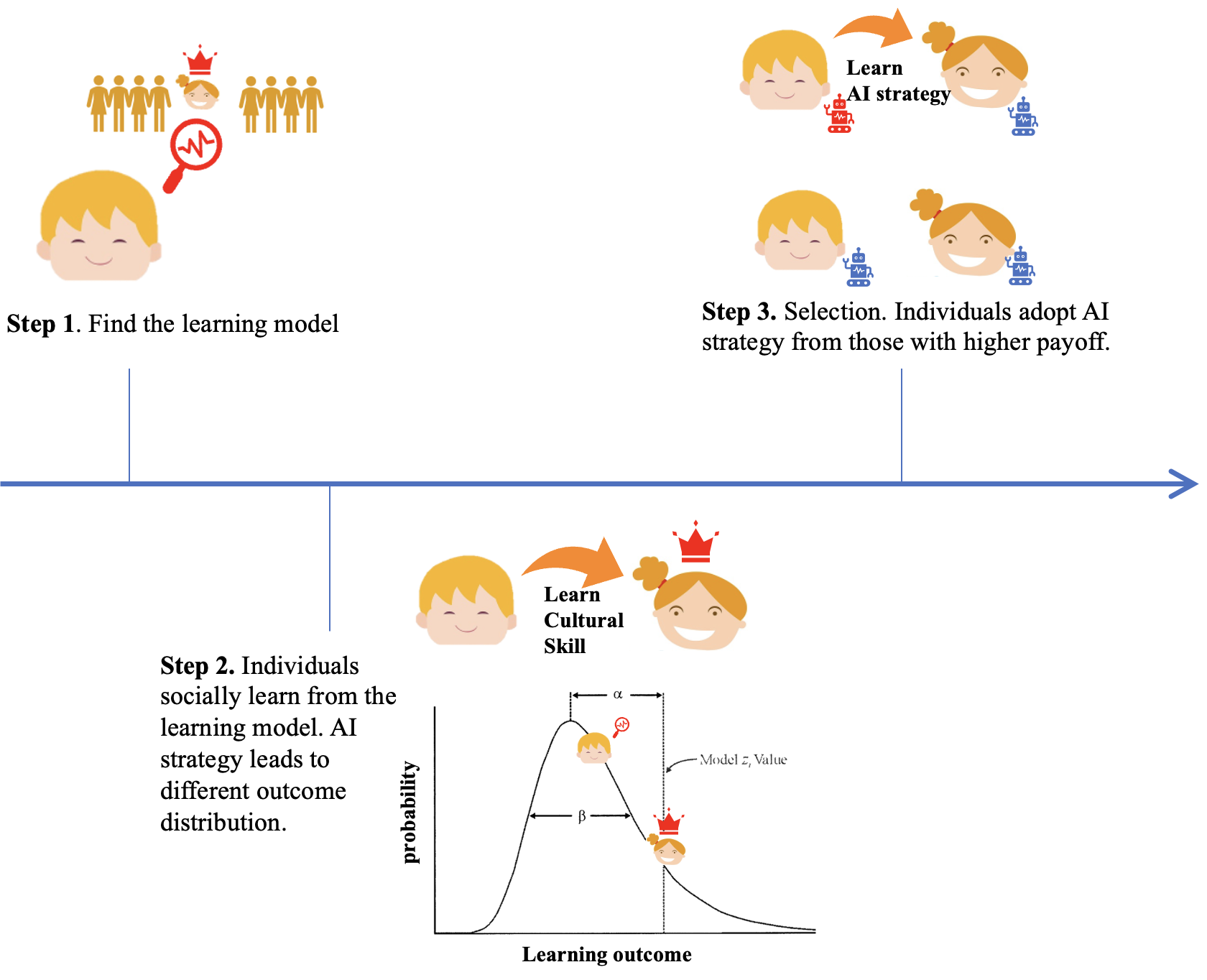}
\caption{\textbf{Model Iteration.} The model runs in three steps iteratively: 1. Searching and learning; 2. Use of AI. 3. Selection on AI strategy.  In a well-mixed population, Step 3 selects for AI strategies based on fitness. With group structure, we assume individuals interact more frequently with in-group than out-group members.}
\label{steps}
\end{figure}

Thus, strategies that tend to yield higher skills (and therefore higher payoffs) are more likely to be adopted and spread. \hl{It is worth noting that we model the learning of the cultural skills and the learning of AI strategy differently. In the CCE process, social learning is more intentional with the goal to maximize cultural skills, so agents preferentially search for the highest-skilled individuals and repeatedly try their best to learn, resulting in a Gumbel distribution. Whereas the learning of AI-use strategies is modeled as a typical spread of innovation with less intention of optimization. Individuals adopt categorical AI strategies through exposure and local interactions. }
A description of the model is presented in Figure~\ref{steps}.

We compare three discrete strategies: no AI, AI Complement, and AI Substitute. This allows us to capture the qualitative effects of categorical AI use on social learning. In additional analyses, we explore the full parameter space of the model.

\section{Results}
\subsection{Effects of AI use on Cumulative Cultural Evolution}

We take two steps to answer how AI will influence cumulative cultural evolution: first, whether the strategy of AI use (no AI vs. AI Complement vs. AI Substitute) will spread in the population, given our assumption about its improvement in accuracy and reduction of variance. Second, once adopted in the population, will the use of AI stall cumulative cultural evolution?

For any particular AI strategy to spread in the population, the AI strategy is required to have a higher average payoff to be adopted through payoff-biased learning. We assume that adopting AI improves learning accuracy. The average payoff is directly related to the improvement of median learning accuracy. Consequently, AI strategies in a mixed population (i.e., some of the population adopting no AI and some adopting any AI strategy) in theory, should be an Evolutionarily Stable Strategy.

\begin{figure}[htp]
\centering
\includegraphics[width=\textwidth]{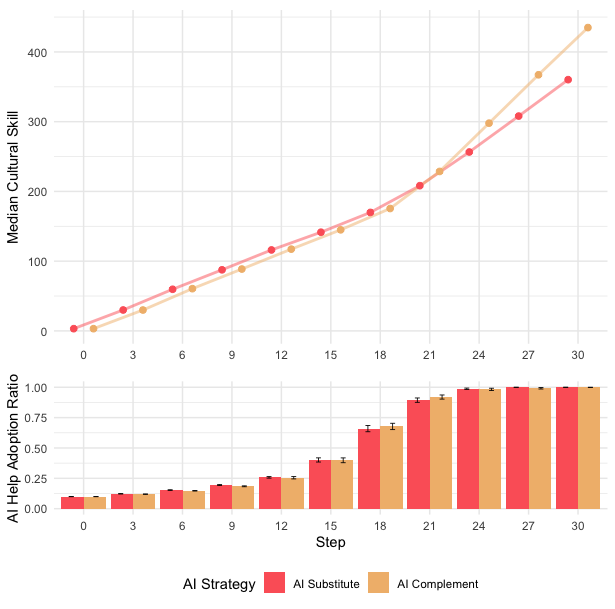}
\caption{\textbf{Cumulative outcome and adoption rate of AI Complement and AI Substitute.} We run the two different conditions (with 100 repetitions) to contrast a population starting with $10 \%$ of AI Complement early adopters and one starting with $10 \%$ AI Substitute early adopters. The line plot in the upper panel shows the median skill across 100 repetitions. Error bars indicating the standard errors between runs are not visible in the figure due to their small size. We observe that even though the AI Substitute condition has a higher median at the beginning due to its immediate high efficiency, the AI Complement condition overtakes it after 18 generations of learning, due to its advantage in variance. The rate of population convergence to the full adoption of AI strategies (lower panel) also influences overall cumulative development. \hl{In Supplementary Figure 4, we show the jump in Step 18 for AI Complement is due to a high maxinum in previous steps, driven by the high variance.}}
\label{cce_independent}
\end{figure}

We first examine the cumulative cultural evolution of individual groups that adopt AI Complement or AI Substitute (RQ1) and focus on within-group dynamics. We illustrate two conditions to show how a quick improvement on collective learning outcomes in the short run can lead to a slower cultural accumulation in the long-run (Figure \ref{cce_independent}): an AI Complement condition with low learning accuracy improvement ($r^{(C)}_\alpha = 0.2$, so $\alpha_C = 0.8\alpha$ )  and low learning variance reduction ($r^{(C)}_\beta = 0.05$, so $\beta_C = 0.95\beta$) in comparison with an AI Substitute condition with high learning accuracy improvement ($r^{(S)}_\alpha = 0.5$) and high learning variance reduction ($r^{(S)}_\beta = 0.5$). We present the evolutionary dynamics of the two conditions, showing that populations have a higher average performance in the long run and converge marginally faster when they adopt AI Complement strategy. See Supplementary Figure 1 for the dynamics and distribution of a single run \hl{and Supplementary Figure 2 for a sweep varying social learning accuracy improvement and variance reduction}. \hl{Additionally, we show in Supplementary Section 3 the conditions where no-AI or AI Complement strategy is the evolutionary adaptive strategy in the long run.} \hl{We also explore the moderation effects of population size on the AI conditions in Supplementary Figure 4. We show that although there is no qualitative difference across different population size,  AI Substitute can dampen the cumulative effects more in a larger population.}

\subsection{Individual selection via replicator dynamics}
\label{sec:replicator}

In addition to the agent-based model, we study individual-level selection on AI strategies using replicator dynamics in a well-mixed population. The objective is to estimate the selection gradient for any population composition of strategies, using the expected payoffs implied by the micro-level searching and learning process described above.

We consider three strategies: no AI use (\(0\)), AI Complement (\(C\)), and AI Substitute (\(S\)). Let
\[
\mathbf{x}=(x_0,x_C,x_S)\in \Delta^2,\qquad x_0+x_C+x_S=1,
\]
denote the population composition, where \(x_s\) is the frequency of strategy \(s\in\{0,C,S\}\).

We take skill to be the payoff-relevant quantity. Specifically, an individual's payoff equals their post-learning skill:
\[
w_i \equiv z_i'.
\]
Given a population composition \(\mathbf{x}\), let \(\pi_s(\mathbf{x})\) denote the expected payoff to a randomly chosen individual using strategy \(s\):
\[
\pi_s(\mathbf{x}) \equiv \mathbb{E}\!\left[z_i' \mid s,\mathbf{x}\right].
\]

Under strategy \(s\), the learning parameters \((\alpha_s,\beta_s)\) are
\[
(\alpha_s,\beta_s)=
\begin{cases}
(\alpha,\beta), & s=0,\\[4pt]
(\alpha_C,\beta_C)=\big(\alpha(1-r^{(C)}_\alpha),\ \beta(1-r^{(C)}_\beta)\big), & s=C,\\[4pt]
(\alpha_S,\beta_S)=\big(\alpha(1-r^{(S)}_\alpha),\ \beta(1-r^{(S)}_\beta)\big), & s=S,
\end{cases}
\]
with \(r^{(S)}_\alpha>r^{(C)}_\alpha\) and \(r^{(S)}_\beta>r^{(C)}_\beta\), as described above.

Conditional on the highest-skilled available cultural model having skill \(z_j\), an individual using strategy \(s\) draws their post-learning skill from the same extreme-value learning distribution as in the agent-based model:
\[
z_i' \sim \mathrm{Gumbel}\left(\text{mode}=z_j-\alpha_s,\ \text{dispersion}=\beta_s\right).
\]
Because the identity and skill of the highest-skilled model \(j\) depend on the population state and the (endogenous) distribution of skills in the population, the expected payoff \(\pi_s(\mathbf{x})\) is generally frequency-dependent.

We estimate \(\pi_s(\mathbf{x})\) via Monte Carlo simulation of the learning step, holding the population composition fixed at \(\mathbf{x}\). For each \(\mathbf{x}\), we proceed as follows:

\begin{enumerate}
    \item \textbf{Initialize a population.} Create \(N\) individuals and assign strategies so that the fractions match \(\mathbf{x}\). Initialize skills \(z_i\) using the same initialization as in the agent-based model.
    \item \textbf{Searching.} In a well-mixed population, each learner identifies the highest-skilled individual \(j\) in the population.
    \item \textbf{Learning.} Each individual \(i\) draws \(z_i'\) from the Gumbel distribution implied by their strategy \(s_i\), with mode \(z_j-\alpha_{s_i}\) and dispersion \(\beta_{s_i}\).
    \item \textbf{Compute payoffs by strategy.} Since payoff equals skill, \(w_i=z_i'\). For each strategy \(s\), compute the average payoff in the population:
    \[
    \widehat{\pi}_s^{(r)}(\mathbf{x})=\frac{1}{N_s}\sum_{i: s_i=s} z_i',\qquad N_s=\#\{i:s_i=s\}.
    \]
\end{enumerate}

We repeat this procedure independently for \(r=1,\dots,R\) replicates (with \(R\) chosen large enough that Monte Carlo error is negligible), and average to obtain:
\[
\widehat{\pi}_s(\mathbf{x})=\frac{1}{R}\sum_{r=1}^R \widehat{\pi}_s^{(r)}(\mathbf{x}) \approx \pi_s(\mathbf{x}).
\]
The resulting mapping \(\mathbf{x}\mapsto(\widehat{\pi}_0(\mathbf{x}),\widehat{\pi}_C(\mathbf{x}),\widehat{\pi}_S(\mathbf{x}))\) summarizes the expected payoff (skill) of each strategy at any population composition.

We treat the expected payoffs as fitnesses in a deterministic selection model. Let the mean payoff be
\[
\bar{\pi}(\mathbf{x})=\sum_{s\in\{0,C,S\}} x_s\,\pi_s(\mathbf{x}).
\]
The replicator dynamics are
\[
\dot{x}_s = x_s\left(\pi_s(\mathbf{x})-\bar{\pi}(\mathbf{x})\right),\qquad s\in\{0,C,S\}.
\]
In practice, we substitute the simulation-estimated payoffs \(\widehat{\pi}_s(\mathbf{x})\) to obtain an estimated selection field on the simplex:
\[
\dot{x}_s \approx x_s\left(\widehat{\pi}_s(\mathbf{x})-\sum_{r\in\{0,C,S\}} x_r\,\widehat{\pi}_r(\mathbf{x})\right).
\]

\begin{figure}
    \centering
    \includegraphics[width=0.8\linewidth]{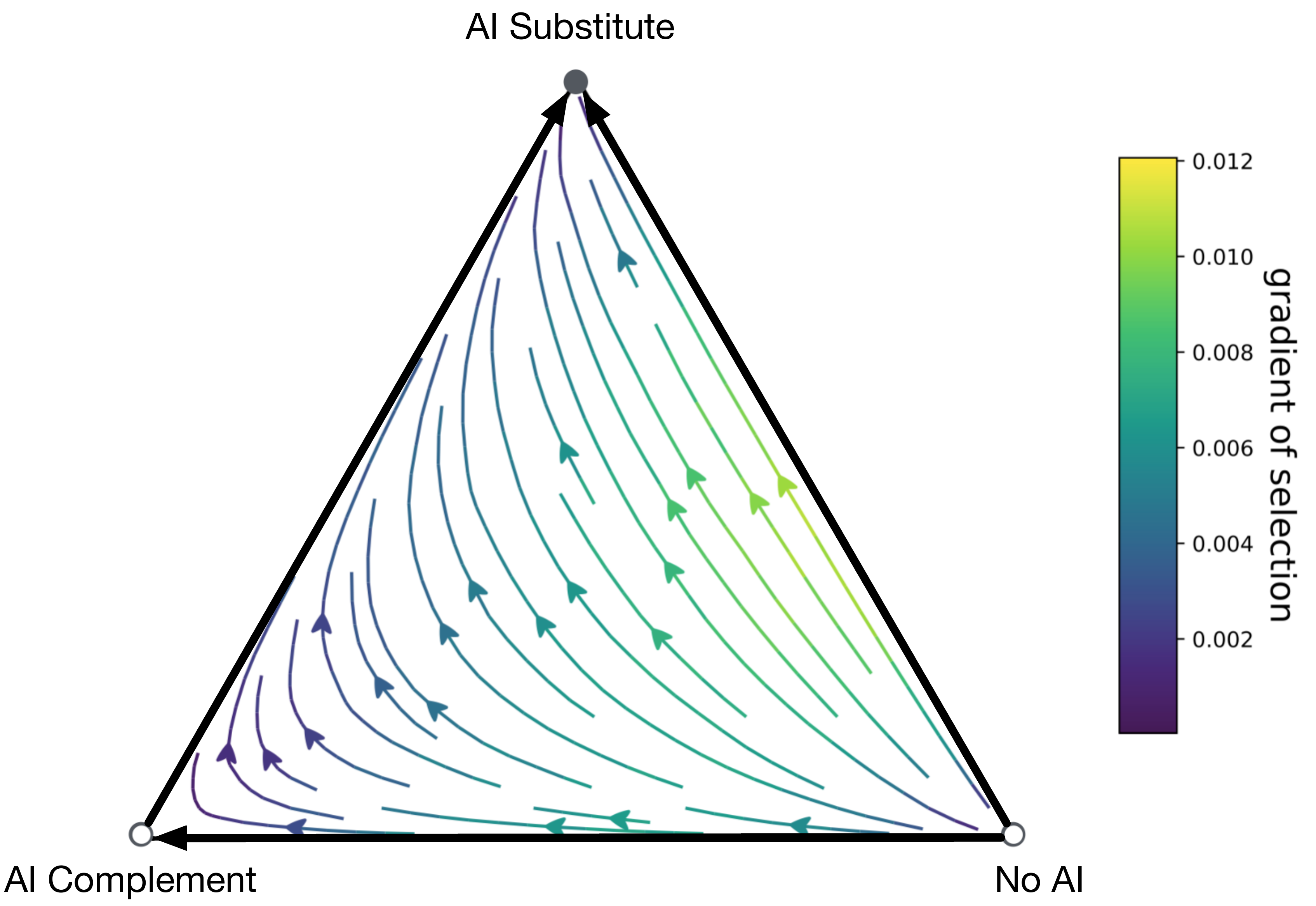}
    \caption{\textbf{Selection gradient on the simplex under replicator dynamics.} 
We plot the replicator vector field on the simplex of strategy frequencies. Payoffs \(\pi_s(\mathbf{x})\) are estimated from repeated simulations of the searching-and-learning process in a well-mixed population and substituted into the replicator equation \(\dot{x}_s=x_s\big(\pi_s(\mathbf{x})-\bar{\pi}(\mathbf{x})\big)\). 
Arrows and representative trajectories show the direction of evolutionary change from different initial compositions; filled points indicate equilibrium compositions. The background colour encodes the speed of selection. 
Along the edges, both AI strategies are favored in pairwise competition against no AI, but AI Complement does not resist invasion by AI Substitute. Consequently, trajectories converge to the AI Substitute, implying this is the unique evolutionarily stable strategy in under standard parameters.}

    \label{fig:simplex}
\end{figure}

This system defines the selection gradient for any population composition \(\mathbf{x}\). Our results are shown in Figure~\ref{fig:simplex}. Under individual selection in well-mixed populations, AI Substitute is the only strategy that survives evolutionary competition. Different cultural learning parameters change the speed of selection, but not the structural outcome. 

\section{Cultural group selection can rescue AI Complement}
Although the risk associated with using GenAI as a Substitute in creative work might be intuitive, the dilemma is that, for most people, using GenAI as a Substitute is more efficient at improving their performance. Yet when everyone uses AI Substitutes, it will lead to the reduction of collective variance and slow down the group's cumulative cultural evolution. How can we create an environment that encourages AI as a complement rather than a substitute, even though it might be more costly and less efficient for individuals in the short run?
 
Cultural evolutionary theory provides evolutionary mechanisms, such as multi-level selection, to address this dilemma between individuals and the group, or short-term and long-term benefits.  The cultural evolutionary approach suggests that group-beneficial and sustainable norms can emerge even without conscious individual planning (\cite{derex2019causal}) or third-party enforcement (\cite{henrich2006costly}). One way individually costly norms (e.g., altruism) can evolve is through group-level selection. Groups with more altruistic individuals have a higher average fitness than groups with more selfish individuals, and thus can spread their altruistic norms to other groups through competition or payoff-biased social learning (\cite{mcelreath2003shared}). Altruism is a costly trait for individuals, especially when other individuals are selfish. Yet groups with more altruistic individuals lower the cost for being altruistic. Over time, groups with more selfish individuals either learn from successful groups or are taken over by them, adopting group-beneficial norms. 

In the context of adopting AI strategies, using AI as a Complement is an individually costly but group-beneficial strategy. To avoid stalling cumulative cultural evolution at the group level, it might be necessary to encourage the use of AI Complements rather than Substitutes. Can we leverage the same evolutionary process of group-level selection to select for AI Complements as the dominant AI strategy?

\textbf{Modeling Multi-level selection through social learning} We extend the model by adding the group-selection mechanism. When there is a group structure, the spread of the high-payoff AI strategy can be altered by the group boundary, allowing for group-level selection. To simplify the behavioral principles of the agents, we follow the example of the spread of coordination behaviors (\cite{mcelreath2003shared}) and use between-group learning to model the process of multi-level selection. 

The differential average skills between different AI conditions allow for multi-level selection. Adopting AI as a Substitute can help individuals get comparative advantages within the group, but reduce the cumulative advantage at the group level.
We implemented a simple group structure where a population of size 1000 is assigned to three groups with equal probability. One group has no AI use in the initial condition. One has $10 \%$ of early adopters of AI Complement. The last one has $10 \%$ early adopters of AI Substitute.

The searching and learning mechanism remains the same as the main model within each group. There is no other structure within the groups. However, to operate group-level selection, we allow agents to adopt AI strategies both within and between groups. 

Because the groups have boundaries between each other, it is easier for agents to interact with other agents within their own groups (with probability $G_1$) than to interact with those from any group (with probability $G_2$, $G_1 + G_2 = 1$, $G_1>>G_2$). \hl{We use the within-group interaction probability $G_1$to operationalize the group boundary. When $G_1 = 1$, groups are isolated from each other and only within group interaction is possible. When $G_1 = 0$, there is no group structure in the population and agents can interact with anyone.} When Agent $i$ encounters another Agent $k$, regardless of within- or between-group interaction, they follow the same strategy learning probability $Pr(i \ copies \ k)$. 

In this way, AI strategies spread faster within the group and slower between the groups. The selection operates at both the group level and the individual level because an individual's learning outcome depends on whether the group can produce a good learning model for the next generation (agents in the next timestep) to learn from and whether the majority are using the appropriate AI strategy that allows for cumulative cultural evolution in the long run.

\subsection{Conditions for multi-level selection}
\begin{figure}[htp]
\centering
\includegraphics[width=\textwidth]{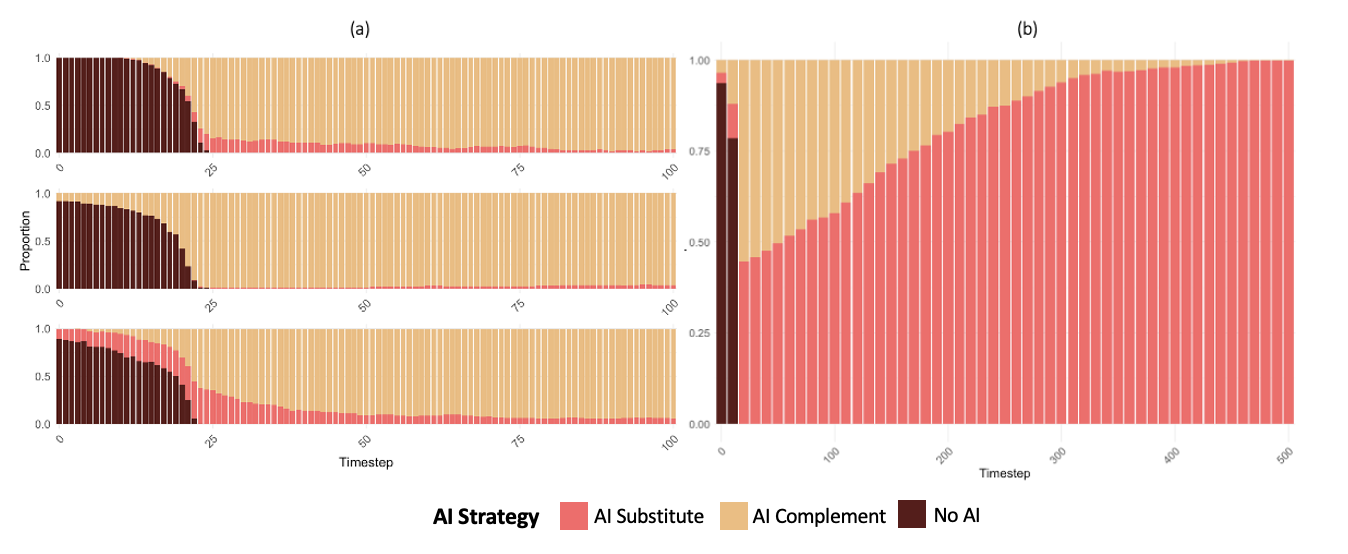}
\caption{\textbf{AI Complement is selected for through group-level selection but not in individual-level selection}. As a group-beneficial but individually costly strategy, using AI as a Complement can be selected for in group-level selection. We run two simulations with the same parameters except for the group structure and in-group learning rate. Panel (a) is one simulation of a population of 1000 in 3 groups. Each strip in this panel shows the AI strategy use in the group across time. We can see that AI Complement (orange) slowly spreads across groups through social learning to become the dominant strategy. Panel (b) is one simulation of a mixed population of 1000. With the same social learning setup, but no group structure, AI Substitute becomes the dominant strategy. In other words, the absence of group-level selection leads to the invasion of individual-beneficial strategies that provide higher short-term benefit. }
\label{multi1}
\end{figure}

As a result, we show that AI Complement can spread as the dominant strategy in the population with the group structure ($G_1 = 0.85, G_2 =0.15$, Figure \ref{multi1}a). We compare this result to the same simulation without the group structure, where individuals can interact with anyone in the population ($G_1=0, G_2=1$). There, the use of AI Substitute spreads among the whole population and becomes the dominant AI use strategy (Figure \ref{multi1}b). The high average learning accuracy $\alpha$ and low learning variance $\beta$ will lead to the highest short-term learning outcome. Therefore, AI Substitute spreads fast in the whole population as the dominant strategy in a mixed population. In contrast, when there is a group structure, groups starting with AI Substitute early adopters will be outperformed by groups starting with AI Complement early adopters in the long run, due to the variance loss. The group boundary slows down the short-term invasion of AI Substitute and protects the variance in the AI Complement group. In the long run, this variance can lead to more cumulative advancement in this group, which drives AI Complement to be the dominant strategy across groups. \hl{We demonstrate in Supplementary Figure 5 this positive causal relationship between social learning variance and cumulative advancement. } 

\begin{figure}[htp]
\centering
\includegraphics[width=\textwidth]{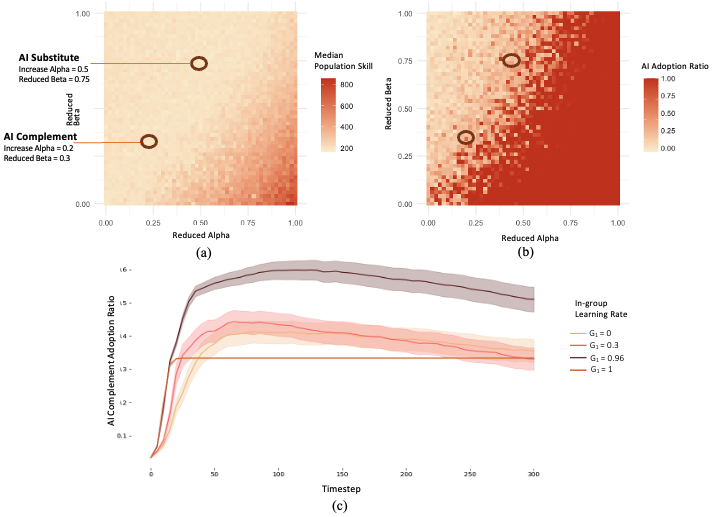}
\caption{\textbf{Conditions for Group Selection.} The differential payoffs enable group-level selection for the group-beneficial AI strategy. Panel (a) shows the median skill level of the population in a single run, varying by the level of AI influence on learning accuracy $D\alpha$ and learning variance $D\beta$. We zoom in to pick two points with the same trade-off ratio $D\alpha/D\beta$ as an example for AI Complement and AI Substitute ($r^{(C)}_\alpha = 0.2, r^{(S)}_\alpha = 0.3, r^{(C)}_\beta = 0.5, r^{(S)}_\beta = 0.75$). Panel (b) illustrates the rate of AI strategy adoption in the two conditions pointed out in Panel (a), with no strong differences ($p_{Complement} = .482, p_{Substitute} = .461$). Although visually there is no strong contrast between the two points, we illustrate in Panel (c) that if we vary the in-group learning rate of the 3 AI strategies (no AI, AI Complement, and AI Substitute), where 0 stands for a mixed population where everyone can learn from everyone else, and 1 stands for a strict group boundary where individuals only learn from in-group members, AI Complement strategy only spreads across the population when the group boundary is relatively strong (above 0.9). At lower in-group learning conditions, we also see a rise in AI Complement adoption due to the stochasticity in the simulation, but it is later invaded by AI Substitute. We present the modeling outcome of different in-group learning ratio for 100 runs. }
\label{comparison}
\end{figure}

After establishing this example of AI Complement and AI Substitute in their specific parameter setting ($r^{(C)}_\alpha = 0.2, r^{(S)}_\alpha = 0.2, r^{(C)}_\beta = 0.4, r^{(S)}_\beta = 0.5$), we return to the full perspective of the parameter space (see Figure \ref{comparison}) and explain the three mechanisms that allow multi-level selection to happen in the 3 panels: (a) The differential rate of cumulative cultural evolution due to the alternation of $D\alpha$ and $D\beta$. (b) The adoption of AI driven by payoff-biased strategy learning throughout time. (c) The high in-group learning rate $G_1$.

This allows us to expand the group structure to 10 equal-sized groups in a population and show the same result, where AI Complement can spread in a structured population but not mixed population (Supplementary Figure 2). 

This long-term advantage of AI Complements can only spread when within-group learning is much stronger than between-group learning. This is because AI Substitutes provide a higher average payoff than AI Complements in each generation. For groups that have early adopters of AI Complement, it requires the adoption of AI Complement to spread within the group soon enough to accumulate the benefit of having more variance than AI Substitute. The group boundary helps slow down the invasion of AI Substitute and protects the variance within the AI Complement group. We show the influence of ingroup learning ratio on the spread of AI Complement in Figure \ref{comparison}.

\section{Discussion}
In this paper, we used an agent-based model to explore the potential effects of the use of GenAI on cumulative cultural evolution. Our results \hl{illustrated a case where} using AI as substitutes \hl{could potentially} hurt our long-term cumulative advantage, even if it can lead to higher efficiency in the short run. At the same time, we also show that group-level selection can function as an evolutionary mechanism that selects for an AI strategy that is beneficial for cumulative cultural evolution. This is the case when the group boundary is strong and light use of AI does not suffer much from the loss of human variance.

\subsection{The long-term and collective effects of AI use}
We presented an abstract agent-based model that can be applied to several cultural practices and productions corresponding to different levels of cultural complexity (represented by $\alpha/\beta$ ). Applications of automation and AI have increasingly shifted from repetitive, low-complexity work (e.g., customer service, email drafting) to creative, high-complexity work (e.g., music production, visual art). The concerns have also shifted from whether AI can excel at creative, complex work to whether AI will replace humans in these areas. 
We study this issue through the lens of cultural evolution, focusing not only on individual performance but on the long-term collective dynamics that emerge when AI is integrated into human learning and production. A key concern is that when AI tools increase efficiency and improve short-term payoffs, they may disproportionately crowd out slower, exploratory human behaviors. This dynamic, we argue, may reduce the collective variance necessary for innovation over time.
Consider the current online music landscape as an example. A growing share of streaming platforms in the "ambient" or "lofi" genre is now occupied by AI-generated music (\cite{ashley2024your}). Putting aside its artistic value, such music is often optimized for minimal disruption and has already claimed a significant share of market presence. This is a case of rapid adoption of AI Substitutes, where low-cost, high-efficiency AI outputs can displace a broad spectrum of independently produced, experimental human music by volume. The risk is not just the homogenization of style, but an erosion of motivation for human innovation, particularly if AI-generated content saturates both the supply and recommendation systems.


\subsection{Out-group learning helps maintain human variance in cumulative cultural evolution}

The paper introduces a multi-level selection framework that favors the AI strategies that maintain the variance of human exploration. Our results have some implications for AI policy-making. In general, we advocate for AI use strategies that protect human exploration and diversity, particularly within the culture, technology and innovation industries, where the tension between regulatory constraints and technological advancement have been key (\cite{blind2012influence}). 

We propose that a more robust, long-term approach to AI alignment involves building an environment that preserves organizational boundaries and promotes model diversity. This structural pluralism enables between-group variation, competition, and adaptive learning, ultimately supporting the selection of AI strategies that align with human long-term interest. 

Our findings also suggest an engineering approach to aligning AI with collective interests. First, instead of building generalized models optimized across different fields and environments, it can be more beneficial to train multiple specialized models in different contexts. This approach, increasingly discussed in the literature on specialist vs. generalist models \parencite{shi2023specialist, fan2024device, nori2023can}, can help maintain diversity across AI systems and reduce the risk of homogenization and model collapse. Second, we advocate for greater variation in training datasets, against the current trend toward training foundation models on increasingly overlapping, centralized corpora (\cite{bender2021stochastic,ganguli2022predictability}).  Maintaining this model and data variance not only contributes to robust and adaptive system design (\cite{page2010diversity}), but also allows for the multi-level selection of group beneficial strategies or designs. This  evolutionary approach of AI alignment complements technical solutions with structural and ecological design.

It is also worth noting that, different from the equilibrium analysis of evolutionary game theory, in this model, the payoff not only depends on the individuals' AI strategy, the opponents' strategy and the frequency of the AI strategy, but also on the group's previous adoption of AI strategy. Individuals are not playing against each other. Instead, they are playing along with the group, and their payoff relies on the AI strategy and the maximum payoff of the group. The maximum payoff of the group is a function of the population of different AI strategy adopters, average learning accuracy, and group learning variance. The feedback loop between individual choice of AI strategy and group learning outcome forms a complex system. 


\subsection{How different types of AI alter social learning}
In this paper, we specifically used generative AI as an example to explore the influence of using AI on human social learning. Different AI applications, however, influence social learning in distinct ways (\cite{brinkmann2023machine}). Recommender systems manage and filter information based on users’ behaviors and preferences, leading to selective retention and transmission of traits. By matching users with content aligned to their preferences, these systems shape both content bias (what is shared) and context bias (who shares with whom). Models like AlphaGo, on the other hand, use reinforcement learning within a well-defined fitness landscape, guiding humans toward high-performing strategies and influencing where exploration occurs. Another example is AlphaFold, which operates in a finite but underexplored scientific space, combining deep learning with human theories and empirical knowledge. While it learns both independently and socially from human input, it does not involve cultural transmission across generations—once the problem is solved, it remains solved. Finally, language models like ChatGPT are trained on the probability distribution of human language, altering not only the content of social learning but also its outcomes by reshaping how knowledge is generated, communicated, and interpreted.

These distinctive applications can have variable influences on cumulative cultural evolution. In a defined landscape, usually the goal is to reach the global maximum. This is the area where we can expect human-AI cooperation to reach the global maximum more efficiently. In an open-ended landscape, the goal is to go further and discover new dimensions in the landscape (\cite{winters2019escaping}). This is the area where heavy use of the current AI technology might pose a long-term risk.

\subsection{Limitations}
This agent-based model is based on an abstract mathematical model that lacks strong empirical quantitative validation. Henrich's base model is well-cited, but often used as a reference to cumulative cultural evolution rather than validated directly (\cite{lindenfors2015empirical, derex2020cumulative, aoki2018absence}). Our model extends on this simple model by making reasonable assumptions based on our current understanding of GenAI. On the one hand, we may need more empirical evidence to support the base model and demonstrate that this mechanism accurately represents human cultural cumulation. On the other hand, we also need experimental evidence to show the effect of AI on the collective social learning process. There have been some attempts in the recent literature trying to measure one-shot experimental outcomes (\cite{anderson2024homogenization}) or the current collective consequences via observational data (\cite{sourati2025shrinking}). These empirical results, to some extent, align with and indirectly support the assumption that AI will reduce social learning variance. In the future, we may still need empirical evidence that demonstrates the mechanism of how AI directly influences the social learning process or the long-term effects of AI on the social learning outcome. 

\hl{Additionally, this model does not explicitly distinguish between effects on individual and social learning. While we formalize an iterative social learning process, performance improvements may also reflect individual learning outcome. We model learning outcomes directly as a Gumbel distribution without specifying whether the improved average performance comes from improved understanding of prior learning models (social learning) or better inference of the environment (individual learning). To focus the model on cumulative effects and social learning, we did not model individual learning separately in this paper, which could be a future direction for this model.}

One possibility that we did not model here is that although AI might slow down cumulative cultural evolution by reducing the variance, human iterative learning using AI might operate at a faster rate, and the overall CCE has already reached a very high level before it is stalled. At that point, it is possible new AI technology or a large AI model population will have emerged to resolve the limitation on variance. 

\hl{Another possible extension is the AI's effect on the search range of learning models. Compared to humans, AI could expand the sample size of information search to find the optimal learning models. Although this advantage may reduce the possibilities when the no-AI strategy is more adaptive, it is unlikely to tip the balance between AI complements and AI substitutes fundamentally. This is because the two strategies differ in how they influence human social learning, rather than in the scope of information searching and sampling.}

\hl{In the present framework, we have a fixed rate of AI influence on social learning accuracy and variance.  However, in reality, this influence is likely more dynamic. AI models may initially drive convergence in human behavior, motivating developers to introduce mechanisms (e.g., increasing temperature, human-in-the-loop, train-of-thoughts methods) to restore diversity. Over time, these adjustments may again lead to homogenization, followed by further innovations that introduce diversity. Understanding the feedback loops between AI development and human behavior is an important future direction. Our model also assumes one general AI system. In practice, individuals may interact with multiple AI systems, compare outputs, or combine different tools. Moreover, users differ in their expertise and strategies for using AI. These sources of heterogeneity at the level of AI systems and user expertise could generate richer dynamics in population-level CCE.}

In summary, this paper explored a novel perspective on how AI may reshape cultural evolution, creative processes, and knowledge production, building on the emerging field of machine culture. Given the accelerating pace of AI development, addressing these questions requires integrating computational simulation with empirical research. Only through this combined approach can we begin to anticipate the societal and cultural influence AI may have and make informed decisions on the future we wish to align with.
\medskip

\newpage
\section*{References}
\printbibliography[heading=none]

\end{document}


\maketitle

\section{Supplementary Figures}

\begin{figure}[H]
    \centering
    \includegraphics[width=\linewidth, height=0.5\textheight,keepaspectratio]{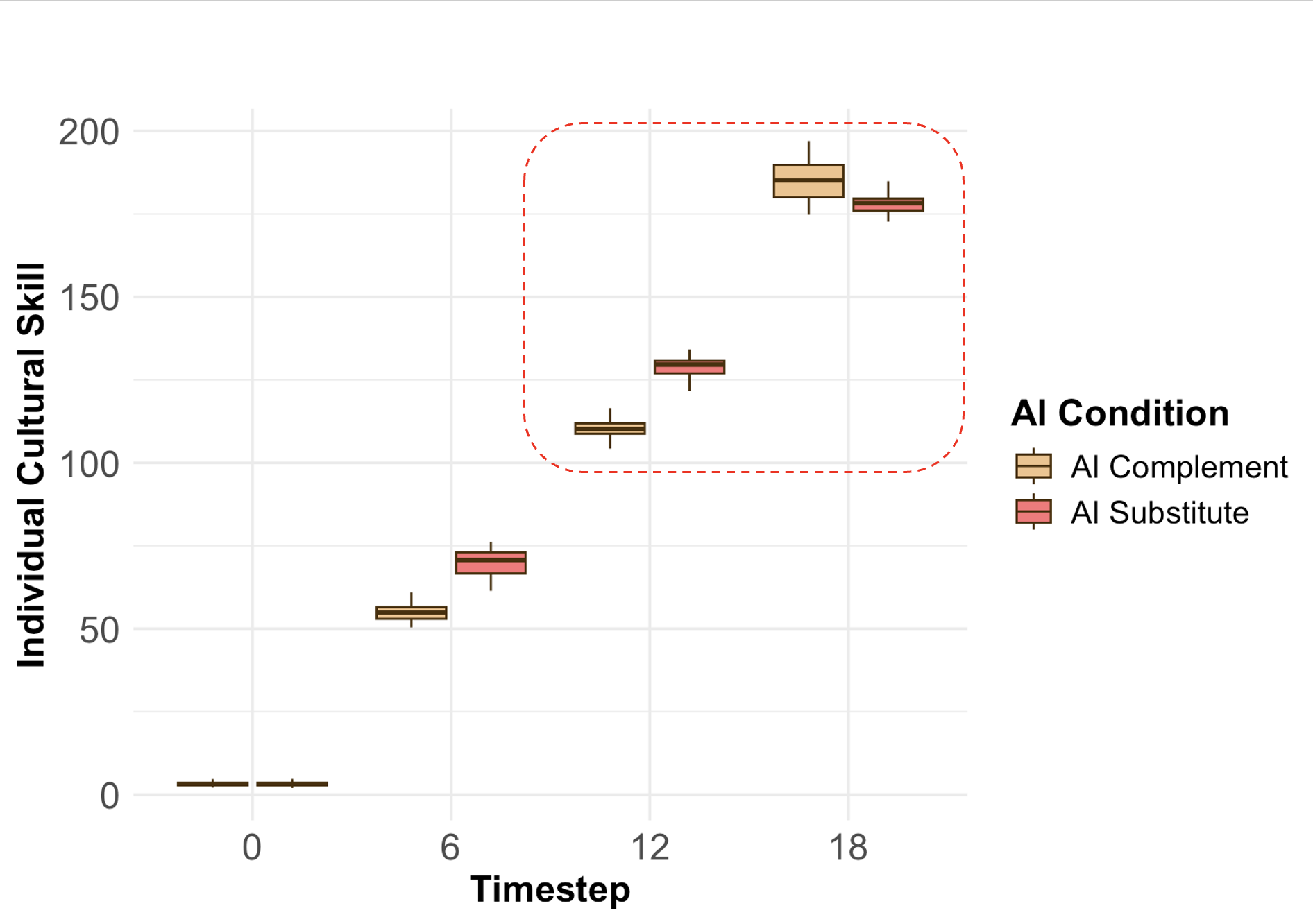}
    \caption{\textbf{Effects of different AI strategies on Cumulative Cultural Evolution}. We compare the results of a single run for the AI Substitute condition and a single run for the AI Complement condition. The box plot zooms into one repetition to show in detail the skill distribution of the two population. In this comparison, we show that even when AI Substitute condition has a higher mean at the beginning due to its immediate high efficiency, AI Complement condition can still overtakes after 18 generations of learning due to its advantage in variance.}
    \label{fig:AI_effect_cce}
\end{figure}

\begin{figure}[H]
    \centering
    \includegraphics[width=\linewidth, height=0.5\textheight,keepaspectratio]{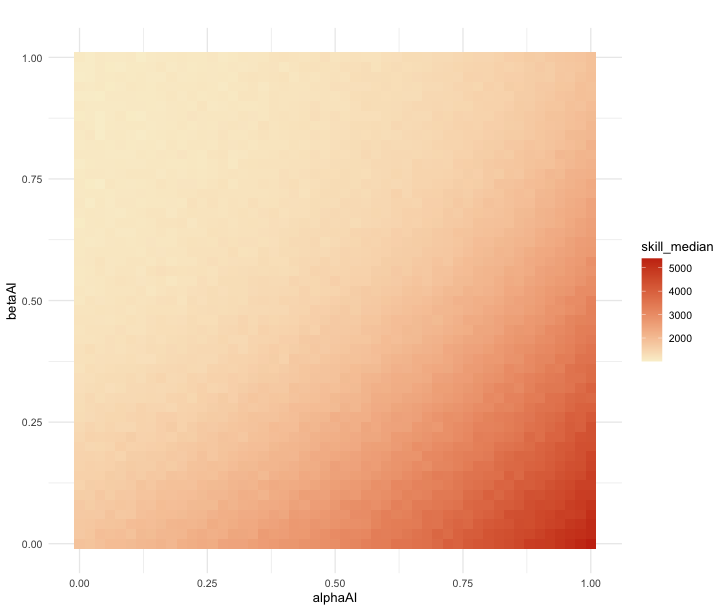}
    \caption{\hl{\textbf{AI Effects of Alpha and Beta on Cumulative Cultural Evolution}. We sweep AI's influence on social learning distance $D\alpha$ and variance $D\beta$ to show its influence on CCE at a more fine-grained perspective. That the more it reduces variance or the less it increases accuracy, the less cumulative evolution it will lead to in 100 run.}}
    \label{fig:sweep_Dalpha}
\end{figure}

\begin{figure}[H]
\centering
\includegraphics[width=\textwidth]{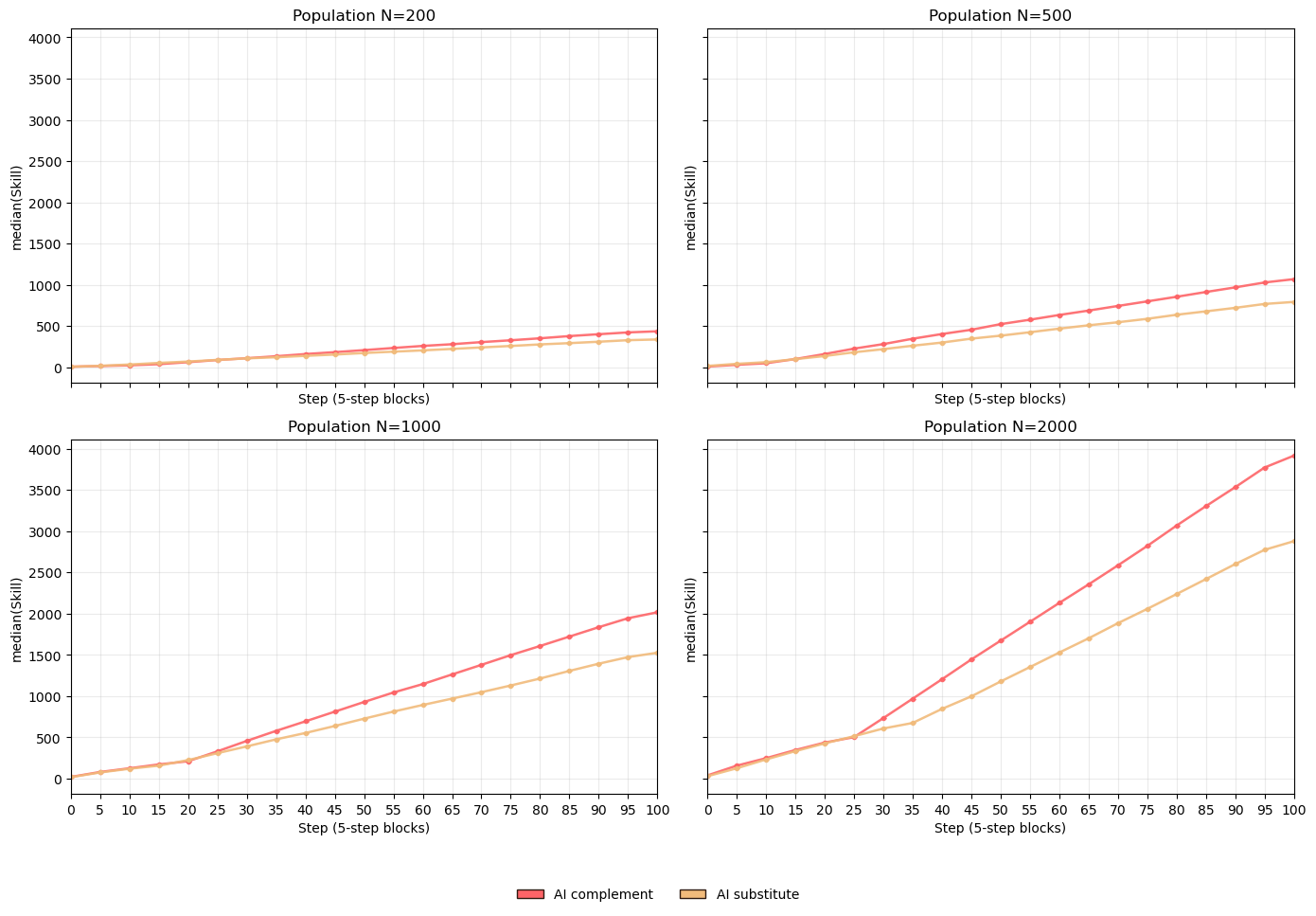}
\caption{\hl{\textbf{AI strategy has a greater influence on large population} We vary the population size to illustrate the influence of AI strategy on different population size. We show that larger population have a higher level of CCE and the cultural skills are more affected by AI strategy. }}
\label{fig:population}
\end{figure}

\begin{figure}[H]
\centering
\includegraphics[width=\textwidth]{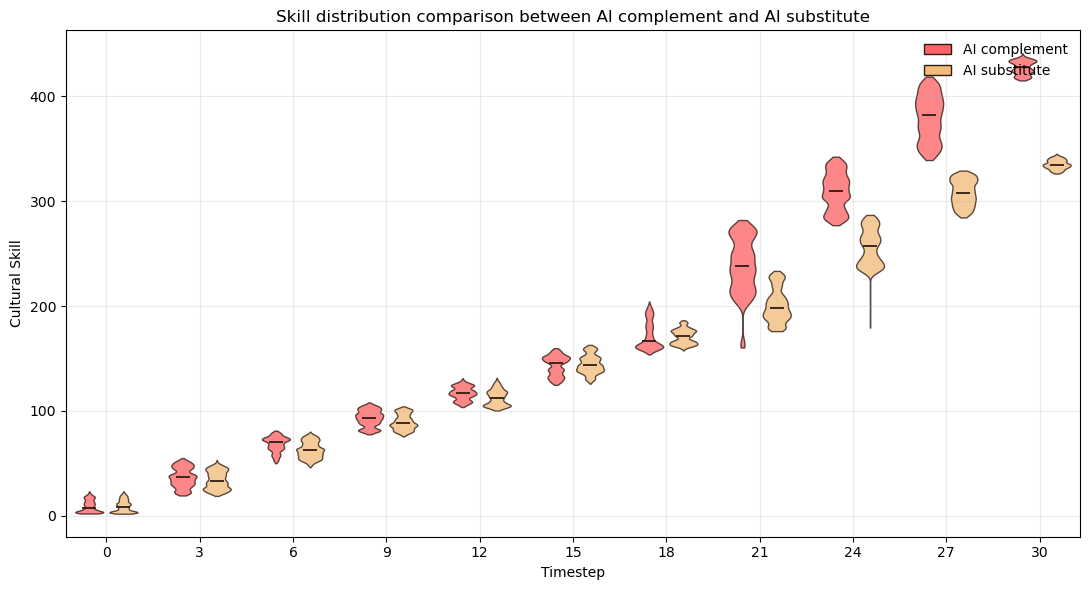}
\caption{\hl{\textbf{Skill distribution of AI conditions} We illustrate the cultural skill distribution to explain why at timestep 18 there is a jump in the AI complement condition. The mean  of AI complement condition at timestep 18 is significantly larger than AI substitute due to the high maxinum in the previous timesteps. In combination with the effect of adoption, this advance in performance and variance leads to a jump in CCE.}}
\label{fig:CCE_distribution}
\end{figure}

\begin{figure}[H]
\centering
\includegraphics[width=\textwidth]{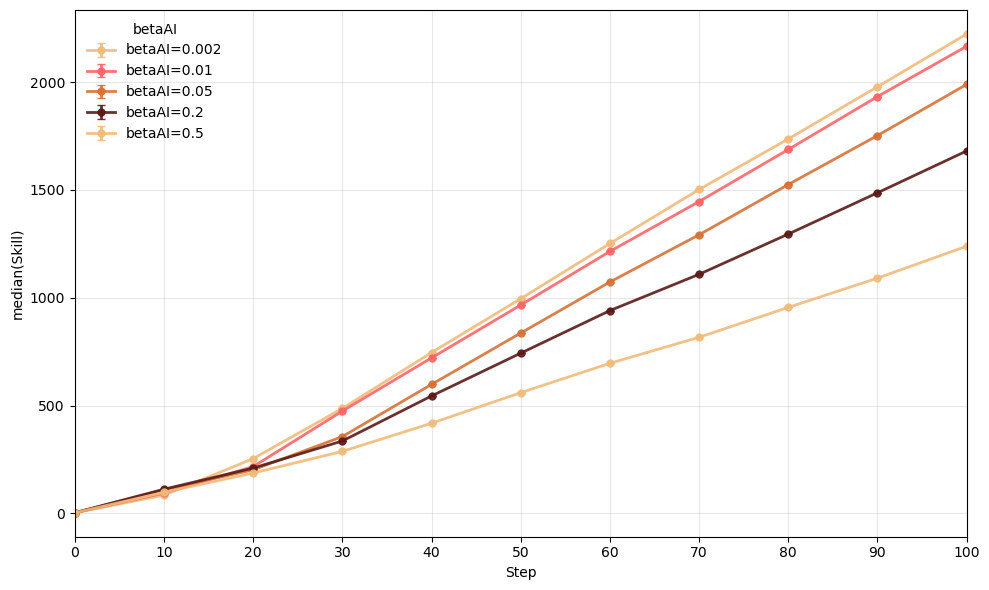}
\caption{\hl{\textbf{Less reduction on variance leads to higher CCE} We vary the AI reduction on social learning variance to demonstrate the causal relationship between social learning variance and the cumulative advancement of cultural skills. This also indicates that it is the social learning variance that allow AI Complements to outperform AI Substitutes.}}
\label{fig:cce_betaAI}
\end{figure}

\begin{figure}[H]
\centering
\includegraphics[width=\textwidth]{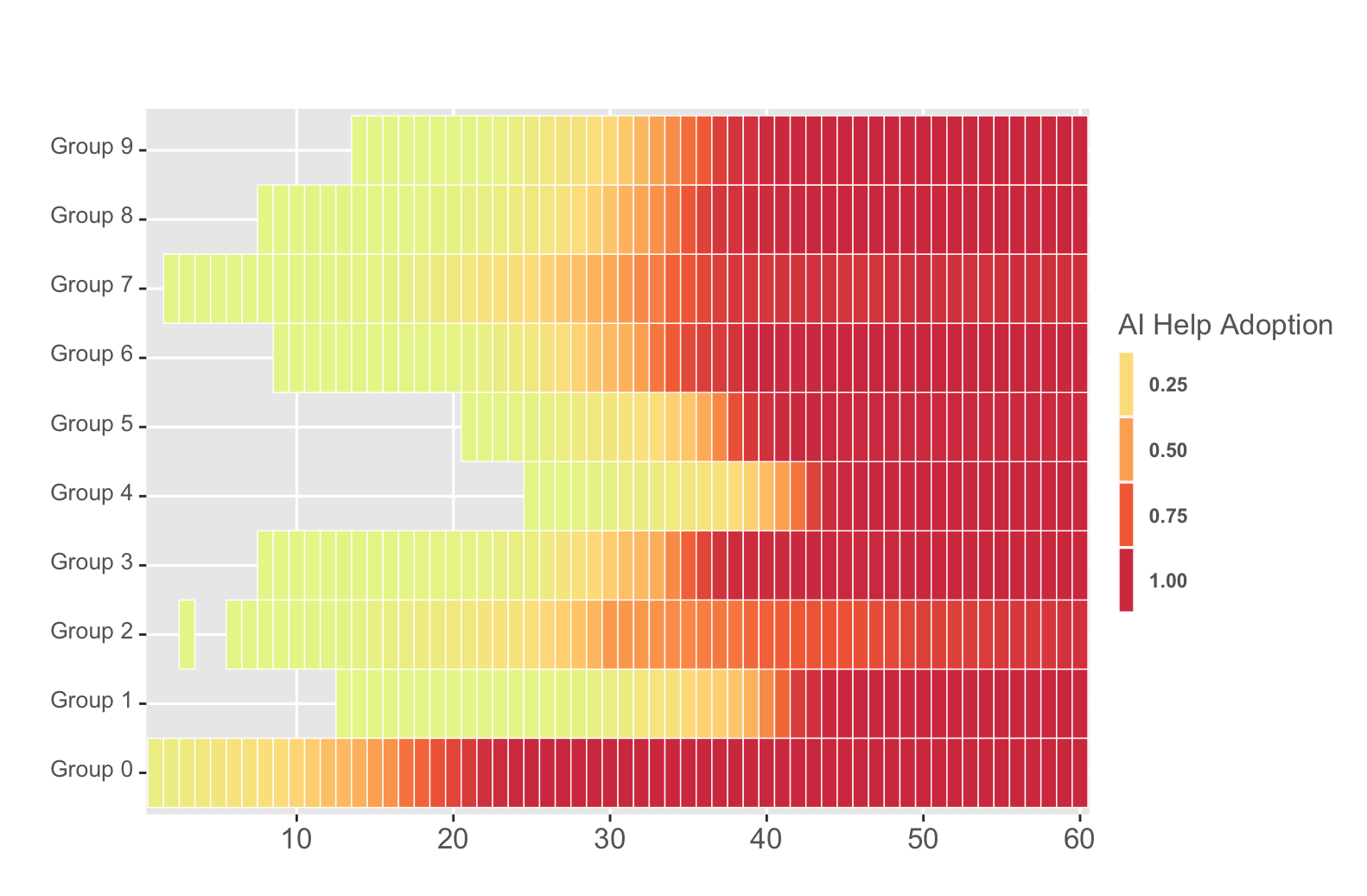}
\caption{\textbf{Multi-level selection of AI Complement across 10 groups} We generalize the group-selection mechanism to 10 groups and show that AI Complement can spread across more groups and become the dominant strategy.}
\label{fig:cce_moregroup}
\end{figure}
\newpage
\section{Model Parameters}
\setlength{\tabcolsep}{10pt} 
\renewcommand{\arraystretch}{1.5}
\begin{tabular}{|p{2cm}|p{4cm}|p{6cm}|}
  \hline
  Experiment & Initialization Parameters & Varied Parameters \\
  \hline
  AI Influence on CCE &  $\alpha = 1$,$\beta = 0.5$, $N = 1000$, $p = 0.1$,$\delta = 10$, $Step = 1000$ & $\alpha = 0.2$, $AI\alpha \sim Uniform(0,1)$, $AI\beta \sim Uniform(0,1)$ \\
  \hline
  Multi-level Selection & $\alpha = 0.2$,$\beta = 0.5$, $N = 1000$, $p = 0.1$,$\delta = 10$, $\alpha = 0.2$, $AI\alpha1 = 0.2$, $AI\beta_1= 0.2$, $AI\alpha_2 = 0.5$, $\delta = 10$, $Step = 1000$  & $AI\beta_2 = 0.5,0.6,0.7,0.8,0.9,0.95$, $G = 0.01,0.1,0.3,0.5,0.7,0.9,0.99$, $m = 1,3,10$ \\
  \hline
\end{tabular}

\vspace{1em}
We carefully pick the initialization parameters so that the system is sensitive enough to the AI modification. For multi-level selection, we picked parameters based on our assumptions that AI replacement can improve more learning accuracy and reduce more variance. We also select parameters in the range where AI Help can be more group beneficial in the long run comparing to AI Replacement to test how varying in-group learning can 

\section{Parameter Space for AI Complement to be the adaptive strategy}
\hl{We show that even though adopting AI Complement might not perform as efficiently as adopting AI Substitute in the short term, it might still be the adaptive strategy in the long run. What is the necessary condition for this to happen? Here we try to explore the parameter spaces where (1) using AI is eventually helpful (or rather, not detrimental) for cumulative cultural evolution; and (2) using AI Complements in the long run is the more adaptive strategy, compared to using AI Substitutes. }

\hl{First, when will the AI strategy stall cumulative cultural evolution? In other words, when is the average payoff of using AI lower than not using AI? We use the terms established in Henrich's model to solve for this parameter space. Our process of CCE happens in a mixed population, where the average payoff depends on the initial rate of adoption and the frequency of AI users at each timestep. To simplify the analysis, here we only compare two populations starting from a different single strategy (e.g., a population always without AI vs. a population always with AI).}

\colorbox{yellow}{In mathematical terms, we need to solve for}
\begin{equation}
 \label{eq:1}
  \Delta \bar{z}_{noAI} > \Delta \bar{z}_{AIstrategy} 
\end{equation}

\hl{Where $ \Delta \bar{z} $ represents the average cultural change of the population across each generation. We follow the previous Henrich model for cultural frequency changes as a function of $\alpha$ and $\beta$ where the approximate cultural change is:}

\begin{equation}
\label{eq:henrich}
    \Delta \bar{z}=p[-\alpha + \beta(\epsilon + Ln(N)]
\end{equation}

\hl{Thus} \ref{eq:1} \hl{can be written as as }
\begin{equation}
\label{eq:3}
-\alpha + \beta(\epsilon + Ln(N)) > -\alpha(1-D\alpha) + \beta(1-D\beta)(\epsilon + Ln(N)) 
\end{equation}

\hl{To find the tipping point for the stall in CCE caused by AI adoption is then to solve for} 
\begin{equation}
    \label{eq:4}
    \frac{D\beta}{D\alpha}>\frac{\alpha}{\beta(\epsilon+Ln(N))}
\end{equation}


\hl{We then explore the parameter space where AI Complement is more adaptive than AI Substitute in terms of cumulative cultural evolution. Using the same structure as Equation }\ref{eq:3}, \hl{we need to solve for} 

\begin{equation}
\label{eq:5a}
 -\alpha(1-D\alpha_C) + \beta(1-D\beta_C)(\epsilon + Ln(N)) >  -\alpha(1-D\alpha_S) + \beta(1-D\beta_S)(\epsilon + Ln(N))
\end{equation}
\hl{We can simplify the inequality to} 
\begin{equation}
\label{eq:5b}
 \frac{D\beta_S-D\beta_C} {D\alpha_S -D\alpha_C}  >  \frac{\alpha} {\beta(\epsilon + Ln(N))}
\end{equation}

\hl{The left-hand side of the equation sketches a trade-off between the extra error deduction and variance deduction. The bigger this term is, the higher the cost of variance deduction is. The right-hand side is a term that incorporates both the cultural complexity and the population that generates enough variance that can afford the variance deduction caused by AI use. For a given cultural complexity ($\alpha/\beta$), the higher the population is, the more space for AI Substitutes to take over, because they can afford more variance loss. }

\begin{figure}[htp]
\centering
\includegraphics[width=\textwidth]{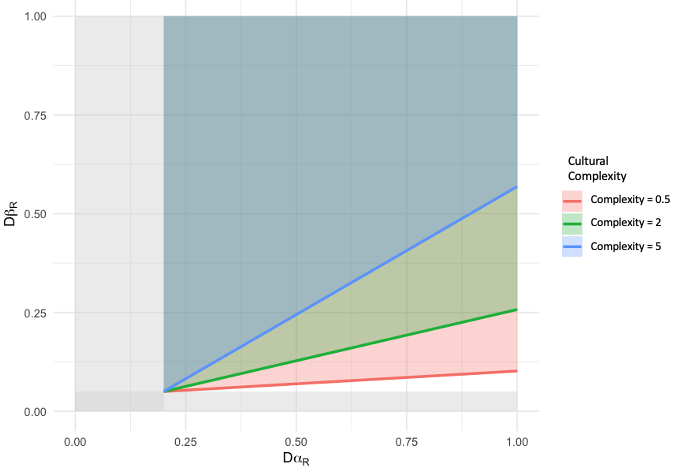}
\caption{\textbf{Parameter Space where AI Complement is the more adaptive strategy.} We solve for $\Delta \bar{z}_{AIComplement} > \Delta \bar{z}_{AISubstitute}$ to get the parameter space where AI Complement outperforms AI Substitute in the long run. To simplify the visualization between parameters, we keep the parameters of AI Complement moderation fixed($D\alpha_S = 0.2, D\alpha_C = 0.05$ ). Given this fixed rate of variance reduction and accuracy improvement of AI Complement, we want to explore the possible variance-accuracy trade-off ratio for AI Substitutes that enables AI Complements to surpass AI Substitutes in long-term cumulative advantage. We plot the range of AI Substitute accuracy improvement $D\alpha_S$ (x-axis) against the range of its variance deduction $D\beta_S$ (y-axis), and show the parameter combination in the colored area where AI Complements have more long-term cumulative advantage and are a more adaptive strategy than AI Substitutes. At the same time, the CCE results also depend on the complexity ($\alpha/\beta$) of the cultural skills people are learning and accumulating. For more complex culture (not easy to copy or vary, e.g., advanced math), there is less likelihood for AI Complement to spread to be the dominant strategy. This is because the advantage of improving learning accuracy exceeds the advantage of keeping a large variance. We plot the cases of different cultural complexity by color and show that the colored areas are smaller with higher cultural complexity, meaning that they require a more narrow range of parameter combinations (more difficult) to satisfy the condition for AI Complement to be the more adaptive strategy.}
\label{tipping_point}
\end{figure}

\hl{We show the parameter space where AI Complement is the more adaptive strategy in Figure} \ref{tipping_point}. \hl{To design a suitable environment or population structure that allows for group selection, we have to look only within the shadowed area, where the long-term benefit of using AI Complement exceeds that of AI Substitute.}

In our current model, the dynamics of population learning are more complicated than this inequality, due to the additional component of AI adoption. Before converging to one type of AI use,  the population has mixed strategies that influence the social learning processes together. These interaction effects between AI adoption and social learning can produce complex dynamics before the whole population converges. 

\hl{To conclude, for the cultural skills that falls in the parameter space in Inequality} \ref{eq:5b}, \hl{we cannot rely on AI to substitute human work. Otherwise we may already face the danger of "cultural collapse". }

